\documentclass[submission,journal]{IEEEtran}
\usepackage{epsfig}
\usepackage{amssymb}
\usepackage{float}
\usepackage{array}
\usepackage[nodots]{numcompress}
\usepackage{graphicx} 
\usepackage{subfigure}
\usepackage{amsmath} 
\usepackage{amssymb}  
\usepackage{cite}
\usepackage{url}
\usepackage{float}
\usepackage{graphicx}
\usepackage[misc]{ifsym}
\usepackage[justification=centering]{caption}

\usepackage[ruled,linesnumbered]{algorithm2e}

\usepackage{booktabs}
\usepackage{multirow}
\usepackage{siunitx}
\usepackage{amsmath}
\usepackage{tabularx}
\usepackage{algpseudocode}
\algnewcommand\algorithmicswitch{\textbf{switch}}
\algnewcommand\algorithmiccase{\textbf{case}}
\algnewcommand\algorithmicassert{\texttt{assert}}
\algnewcommand\Assert[1]{\State \algorithmicassert(#1)}%
\algdef{SE}[SWITCH]{Switch}{EndSwitch}[1]{\algorithmicswitch\ #1\ \algorithmicdo}{\algorithmicend\ \algorithmicswitch}%
\algdef{SE}[CASE]{Case}{EndCase}[1]{\algorithmiccase\ #1}{\algorithmicend\ \algorithmiccase}%
\algtext*{EndSwitch}%
\algtext*{EndCase}%
\newcommand{\tabincell}[2]{\begin{tabular}{@{}#1@{}}#2\end{tabular}}
\newcolumntype{N}{@{}m{0pt}@{}}

\usepackage{diagbox}
\newcolumntype{N}{@{}m{0pt}@{}}

\begin{document}
%
\title{InsClustering: Instantly Clustering LiDAR Range Measures for Autonomous Vehicle }

\author{You Li$^{1}$, Cl\'ement Le Bihan$^2$, Txomin Pourtau$^2$, Thomas Ristorcelli$^2$ 
\thanks{$^{1}$You Li and Javier Ibanez-Guzman are with research department of RENAULT S.A.S, 1 Avenue du Golf, 78280 Guyancourt, France
        {\tt\small you.li@renault.com}}
\thanks{$^{2}$Clement Le Bihan, Txomin Pourtau, Thomas Ristorcelli are with Magellium S.A.S, Parc technologique du Canal,  24 rue Hermès, 31521 Ramonville Saint-Agne Cedex, France 
        {\tt\small \{clement.lebihan, txomin.pourtau, thomas.ristorcelli\}@magellium.fr}}%
}
    
\maketitle

\begin{abstract}
  LiDARs are usually more accurate than cameras in distance measuring. Hence, there is strong interest to apply LiDARs in autonomous driving. Different existing approaches process the rich 3D point clouds for object detection, tracking and recognition. These methods generally require two initial steps: (1) filter points on the ground plane and (2) cluster non-ground points into objects. This paper proposes a field-tested fast 3D point cloud segmentation method for these two steps. Our specially designed algorithms allow instantly process raw LiDAR data packets, which significantly reduce the processing delay. In our tests on Velodyne UltraPuck, a 32 layers spinning LiDAR, the processing delay of clustering all the $360^\circ$ LiDAR measures is less than 1ms. Meanwhile, a coarse-to-fine scheme is applied to ensure the clustering quality. Our field experiments in public roads have shown that the proposed method significantly improves the speed of 3D point cloud clustering whilst maintains good accuracy. 
\end{abstract}

\section{INTRODUCTION}\label{sec::intro}
Perception systems of autonomous vehicles usually consist of cameras, LiDARs and radars \cite{youliSPM2020} \cite{YouLIHDR}. As an active sensor, LiDAR illuminates the surroundings by using laser devices. Distances are measured through analyzing the laser returns. Currently, the most popular LiDARs are mechanical spinning LiDARs (e.g. Velodyne HDL64), which scan a wide scenario through physically rotating lasers and receivers. LiDAR is much better than the camera in the accuracy of distances but poor in object recognition due to the sparsity of point clouds. Radar can work in all-weather conditions and is better in moving object detection and speed estimation. However, the resolution of radar is insufficient for complex environments. Therefore, LiDAR is always regarded as an important sensor to ensure safety for high level autonomous driving in complex environments.

In the perception systems for autonomous vehicles, LiDARs are generally used for object detection, tracking and recognition.  Sometimes, LiDARs are used for \textit{Occupancy grid mapping}\cite{edouardITSC2018} for path planning. \cite{himmel2008} proposed a classic LiDAR perception pipeline to extract obstacles from 3D point cloud: (1) \textit{clustering}: point clouds are firstly separated by ground segmentation. Then, non-ground points are grouped into different obstacles. (2) \textit{object recognition} and \textit{tracking:} obstacles are classified into semantic classes \cite{Alex2011, edIV2019} (e.g. pedestrian, vehicle, truck etc) and tracked through Bayesian filters \cite{Anna2009}.

In this paper, we focus on point cloud clustering, i.e. \textit{ground segmentation} and \textit{object clustering}. As the first step of the whole LiDAR perception pipeline, segmentation is required to be as \textit{fast} and \textit{accurate} as possible. For a Velodyne LiDAR working at 600RPM (rotation per minute), the processing time of the perception algorithms is usually expected to be less than 100ms. Considering that tracking and classification tasks for multiple objects are more complicated and requir more time, the time budget for point cloud segmentation is quite limited.

The proposed approach "InsClustering" is able to instantly process the received raw UDP packets from a popular LiDAR: Velodyne UltraPuck. Range image in spherical coordinates is utilized to process the raw packets for ground segmentation and object clustering. Finally, a coarse-to-fine scheme is applied to maintain the clustering accuracy, especially mitigate the over-segmentation problem. The main contributions of this method are:
\begin{itemize}
\item A fast 3D point cloud segmentation for autonomous driving scenarios. The time delay of completing the segmentation of a full $360^\circ$ scan of a 32 layers spinning LiDAR is less than $1ms$ on our platform (Intel i7-7820, 16G Memory, not rely on GPU), which is faster than two SOTA methods DepthClustering \cite{Igor2016}, and SLR \cite{Dim2017}.
\item Good accuracy performance, especially in ameliorating over-segmentation problem, compared to the two SOTA methods. 
\end{itemize}
  
\section{Literature Review}\label{sec::review}
As the first step of applying LiDAR into autonomous driving, the ground segmentation and object clustering have been well researched. The related methods can be roughly divided into:  

\textit{2.5D grids based methods}: By projecting 3D points into 2D $xy$ plane, 2.5D grids (plus height information) are created. In the early works of \cite{Anna2009, PerceptionJFR2008}, the number of points, maximum height within a 2.5D polar grid and angles between consecutive grids are used for ground segmentation. The non-ground points are grouped through a chunk clustering method in \cite{PerceptionJFR2008}. The same 2.5D polar grids structure is kept in \cite{himmel2008}, but line segments are fitted to judge whether a grid belongs to ground or not. The left non-ground grids are clustered by the connected component labeling (CCL). 

\textit{Spherical coordinates (range image) based methods:} Instead of projecting onto a 2D ground plan, 3D points can be viewed through Spherical coordinates $(r,\varphi,\theta)$ according to characteristics of scanning LiDAR. For mechanical scanning LiDAR, vertical angle for each laser beam is fixed, azimuth angle is decided by the scanning time and motor speed. Therefore, every range reading can be represented by $P_{i,j} = (\rho_{i,j}, \varphi_i, \theta_j,)$, where $i$ refers to a certain laser beam, $j$ is the azimuth angle index. Avoiding the creation of 2.5D polar grids, this approach naturally fills the range readings into a predefined range image data buffer and allows quick access to a point and its neighbours. For example, \cite{Igor2016} firstly segments the ground points in each column of a range image, and group the remaining points by criterions of distance and angle. For 32 beams LiDAR, they reached 4ms in an Intel i5 U5200 2.2 GHz CPU, C++ implementation under ROS. The ground segmentation in \cite{Dim2017} is realized based on 2.5D grids, while for the object clustering, they firstly apply the clustering in each scan-line (actually the row in range image), and then merge the clusters scan-line by scan-line. 

\textit{Graph based methods:} this is more general than 2.5D grids and range image, usually coupled with graph based algorithms. For instance, \cite{Moosmann2009} represents 3D point clouds by an undirect graph. Local plane normals are compared to cut the edges of graph to realize ground segmentation. However, it is not a real-time method. \cite{Douillard2011} proposed two different segmentation methods, Gaussian Process Incremental Sample Consensus (GP-INSAC) and a mesh-based method inspired by \cite{Moosmann2009}. However, these two methods are not real-time. \cite{lukas2017} models the ground as a spatio-temporal conditional random field, dividing the surrounding into connected grids. Heights of each grid are estimated concurrently by an EM variant algorithm implemented for GPU. Although the results are promising, this method is not real time. 

\textit{Deep learning based semantic segmentation:} In \cite{PointNet2018}, the authors designed a neural network to semantically segment 3D point clouds. However, due to the sparsity of LiDAR data w.r.t distance, this method doesn't work well for autonomous driving scenarios. \cite{SqueezeSeg} applies convolutional neural networks for semantic segmentation of objects from LiDAR point clouds. This method is capable of real-time processing thanks to the use of the range image. 

\section{Methodology} \label{sec::method}

\subsection{Sensor model}
\begin{figure}[h]
\centering
\subfigure[Spherical coordinates and Velodyne VLP32C]{
\includegraphics[width = 0.4\textwidth]{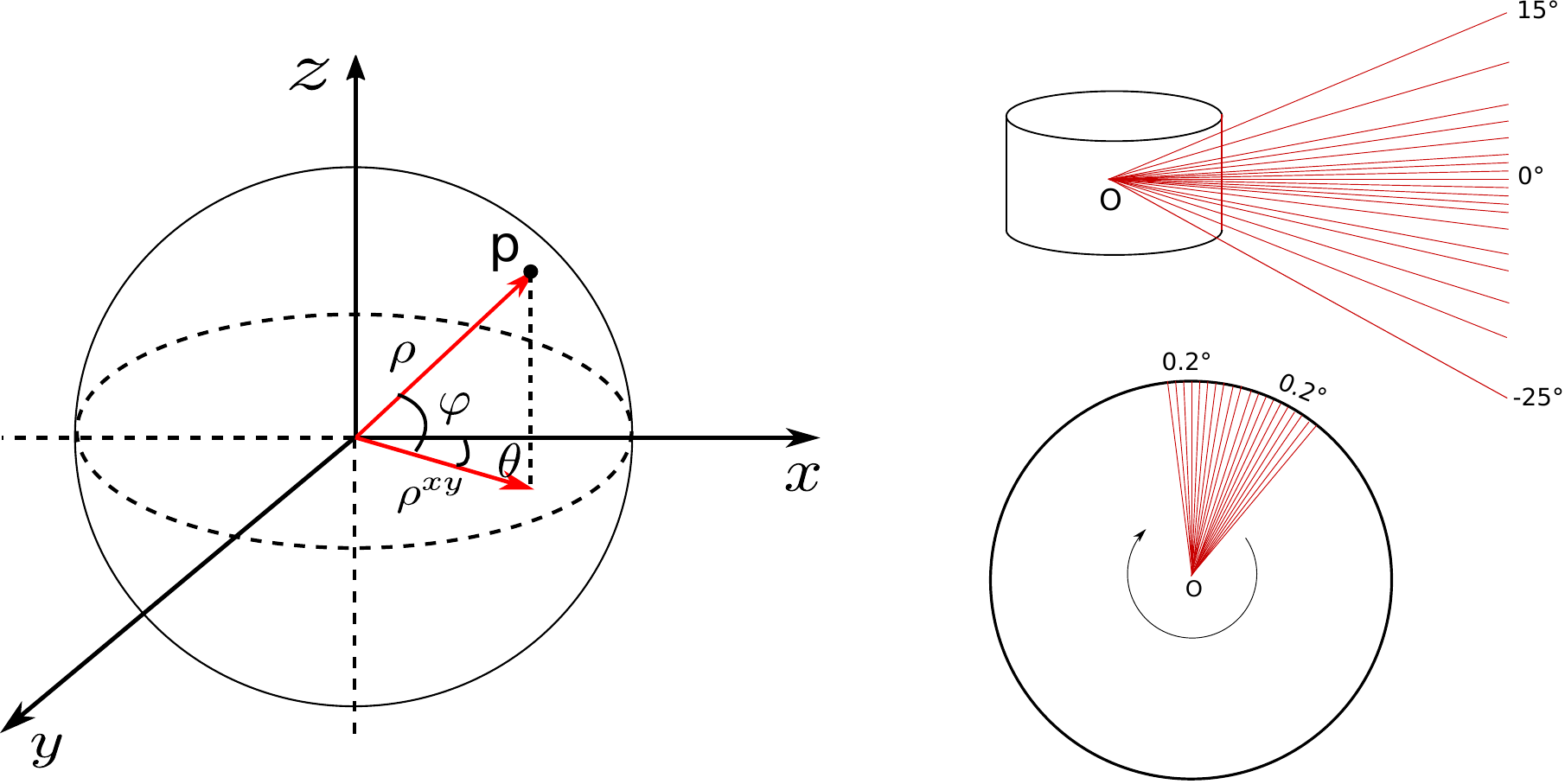}
}
\subfigure[A range image model (origin from bottom left) and a real example (in pseudo color)]{
\includegraphics[width = 0.48\textwidth]{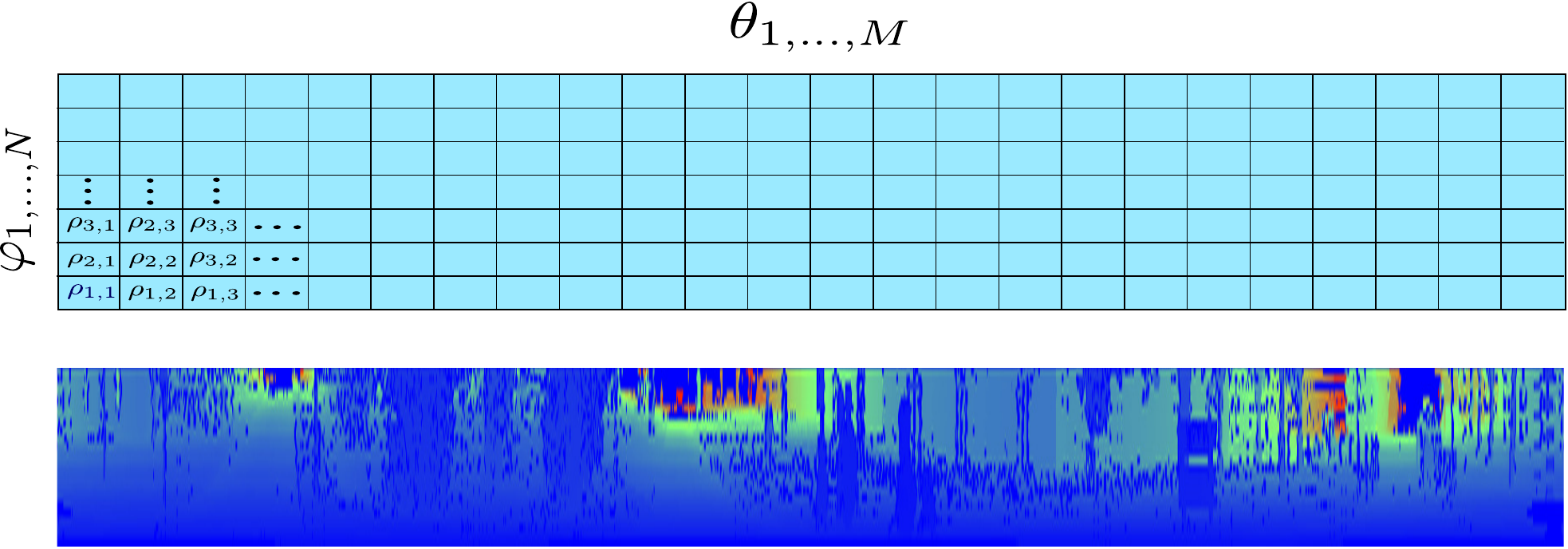}
}g
\caption{\small{Sensor model and range image}}
\label{fig::lidar}
\end{figure}

The proposed method follows the principle of range image, as reviewed in Sec. \ref{sec::review}. All the laser beams fired from a spinning LiDAR can be considered to be emitted from an origin point $O$ in a spherical coordinate system: $(\rho,\varphi,\theta)$ (illustrated in Fig. \ref{fig::lidar} (a)), where $\rho$ refers to range measure, $\varphi$ and $\theta$ are the vertical and azimuth angles respectively. A range image $\mathcal{I_{N\times M}}$ can be easily constructed as shown in Fig. \ref{fig::lidar} (a). We define a point $p$ in range image $\mathcal{I}_{ N\times M}$:  
\begin{equation}
  p_{i,j} = [\rho_{i,j}, \varphi_i, \theta_j]^T \;  i\in[1,N], j\in[1,M]
  \label{eq::range_image}
\end{equation}
Two state-of-the-art methods, SLR \cite{Dim2017} and DepthClustering\cite{Igor2016} both use depth image for point cloud clustering. While SLR process the range image row by row, and DepthClustering is column by column. Fig. \ref{fig::lidar} (b) shows an real example of a range image for the used Velodyne UltraPuck LiDAR.  


 \begin{figure*}[t]
  \centering
\includegraphics[width = 0.65\textwidth, height = 6cm]{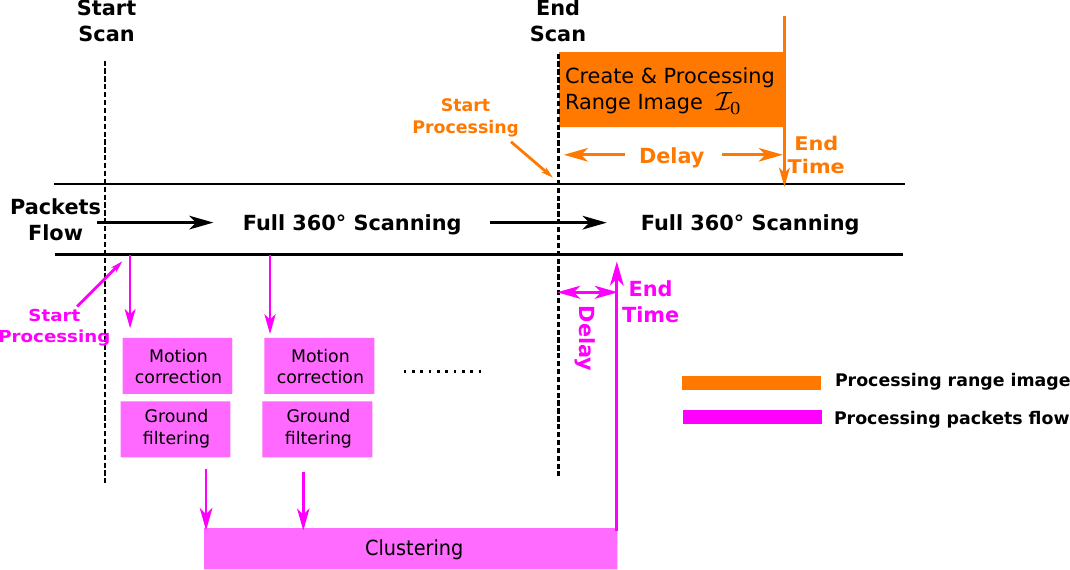}
\caption{Difference of processing range image and raw data flow. The end time of processing of range image based approach is later than the one of packet flow based approach.}
\label{fig::process_packet}
\end{figure*}

\subsection{InsClustering}
\label{sec::rawdata}
The range image as used in SLR\cite{Dim2017} and DepthClustering \cite{Igor2016} provides a nature and memory-friendly representation for processing a spinning LiDAR's outputs. Nevertheless, being different to a global shutter camera that captures an image at once, a spinning LiDAR continuously outputs the range measures during its scanning process. Generating a range image causes the problem of latency:

\textit{Latency problem:} Waiting for a LiDAR to finish a full scan and then processing the range image leads to delay of object detection. In the worst case, when a Velodyne LiDAR is spinning at 600RPM (100ms for $360^\circ$ scan), suppose there is an obstacle that appears at the very beginning of a scan. Then, we will wait almost 100ms for the range image creation and then pay for the algorithm processing time to finally receive the detected obstacle. Therefore, due to this defect of range image solution, a perception system would have late obstacle detections, which is critical for autonomous driving functions, such as autonomous emergency braking (AEB).

To cope with this problem, the proposed method InsClustering adapts traditional range image processing framework to directly process raw data flow from a spinning LiDAR. As denoted by the name, this method can instantly process the received LiDAR data and generate object detections without extra delay. The principle is shown in Fig. \ref{fig::process_packet}.    

\subsubsection{Handle raw data packets}
In single return mode, the utilized Velodyne UltraPuck continuously outputs a stream of data packets $\mathcal{P}^t$ via UDP protocol. The 1248 bytes long data packet $\mathcal{P}$ comprises 12 data blocks, and each data block contains an azimuth angle and 32 range measures corresponding to the 32 lasers fired simultaneously. In fact, $\mathcal{P}$ is a part of range image $\mathcal{I}_{N*M}$. Although in UltraPuck, the 32 lasers are not in the same azimuth angle, $\mathcal{P}$ can still be viewed as several columns of $\mathcal{I}_{N*M}$, after accumulating enough data packets. 

\begin{algorithm}[t]
   \small
  \SetAlgoLined
  \KwIn{Velodyne data packets flow}
  \KwOut{Ground labels and clusters}
\tcc{\small{Step 1: ground segmentation}}
\For{every $\mathcal{B}_{N*M'}^i$ built from data packets}{
  Coarse and fine ground segmentation in $\mathcal{B}_{N*M'}^i$ as step 1 and 2 in Algorithm \ref{alg::ground_segmentation}.\\
}

\tcc{\small{Step 2: Clustering}}
\For{every obstacle point $p\{l^{o}\}$ in received $\mathcal{B}_{N*M'}^t$}{
  Coarse clustering within a searching region $\mathcal{S}(p\{l^{o}\})$, send initial clusters into cluster buffer $\mathcal{CB}$\\
}
Refine the cluster buffer $\mathcal{CB}$ according to Algorithm \ref{alg::clustering}. \\
Output the refined clusters in $\mathcal{CB}$ when necessary. 
\caption{Processing on LiDAR raw data flow}
\label{alg::process_packets}
\end{algorithm}

Instead of processing $\mathcal{I}_{N*M}$, the new approach instantly processes a buffer of received data packets $\mathcal{P}^t$, and thus ameliorates the latency (as shown in Fig. \ref{fig::process_packet}). Let $\mathcal{B}_{N*M'}^i$ be one packet buffer created by a number of sequentially received data packets, where $N$ equals to the number of rows in $\mathcal{I}_{N*M}$, $M'$ is the buffer length. After creating the packet buffer, the proposed algorithm can directly process it. Algorithm. \ref{alg::ground_segmentation} presents the pseudocode of InsClustering.  

Within InsClustering, ground segmentation and a coarse clustering are performed instantly for each received packet buffer. Since the size of obstacle could be bigger than the length of packet buffer, a cluster buffer is maintained to refine previous generated clusters and output completed clustering results.     



\subsubsection{Ground segmentation} \label{sec::ground_segmentation}


A fast coarse segmentation is first performed in each input packets buffer $\mathcal{B}_{N*M'}^i$, which is $M'$ columns of a range image $\mathcal{I_{ N\times M}}$. The algorithm starts from a virtual point at the very beneath of the LiDAR. The virtual point is decided by the vehicle's height and it is always assumed to be on the ground. Then, following the increasing order of vertical angles, the algorithm sequentially accesses point by point. Inspired by \cite{Chu2017a}, any input point is firstly evaluated to be a \textit{change} point or not, based on its \textit{slope}. A \textit{change point} usually reveals a start of an obstacle. For a point $p_{i,j}$, its \textit{slope} $\alpha_{i,j}$ is defined as $(z_{i+1,j}-z_{i,j})/(\rho^{xy}_{i+1,j}-\rho^{xy}_{i,j})$. If $\alpha_{i,j}$ is bigger than a threshold $t_{\alpha}$, $p_{i,j}$ is classified as a \textit{change} point. This is described as $\textit{isChangePoint()}$ in Algorithm \ref{alg::ground_segmentation}.

If $p_{i,j}$ is not a \textit{change} point, its previous class $l_{i-1,j}$ is an important clue for decision. In \cite{Chu2017a}, if a point following the change point has a big height decrease, it will be labeled as \textit{ground} again. Nevertheless, we observe that, due to the occlusion, the points far from a \textit{change} point are hard to be classified by simple height changes. Therefore, a temporary label \textit{uncertain} point is given in this case. $p_{i,j}$'s \textit{range difference} $\Delta\rho^{xy}_{i,j}$ is calculated with respect to its next point $p_{i+1,j}$. If $\Delta\rho^{xy}_{i,j} \geqslant t_{\Delta\rho^{xy}}$, the point will be labeled as \textit{uncertain}. Otherwise, it will be labeled as \textit{change follow}. This is described as function $\textit{isClose()}$ in Algorithm \ref{alg::ground_segmentation}.

\begin{algorithm}
\small
 \SetAlgoLined
  \KwIn{All the points $\textbf{p}$ in one packet buffer $\mathcal{B}_{N*M'}^i$}
\KwOut{Ground points $\textbf{p}\{l^g\}$, obstacle points $\textbf{p}\{l^o\}$}
\tcc{\small{Step 1: Coarse ground segmentation}}
\ForEach {point $p_{i,j}$}{
    \eIf{\textit{$isChangePoint(p_{i,j})$} == true}{
    Label $p_{i,j}$ as \textit{change point}: $l_{i,j} = l^c$  
  }
  {
    \Switch{previous status: $l_{i-1,j}$}
    {
      \Case{$l_{i-1,j}$ is \textit{ground} point}{
        Label $p_{i,j}$ as \textit{ground}: $l_{i,j} = l^g$
      }
      \Case{$l_{i-1,j}$ is \textit{change} or \textit{change follow} point}{
        \eIf{$\textit{isClose}(p_{i-1,j}, p_{i,j})$ == true}{
        Label $p_{i,j}$ as \textit{change follow}: $l_{i,j} = l^{cf}$   
        }
        {
          Label $p_{i,j}$ as \textit{uncertain}: $l_{i,j} = l^u$
        }
      }
      \Case{\textit{uncertain}}
      {
        Label $p_{i,j}$ as \textit{uncertain}: $l_{i,j} = l^u$
      }
    }
  }
}
\tcc{\small{Step 2: Fine ground segmentation}}
Fit line segments $\mathcal{L}_s$ in packet buffer, $s\in[1,S]$\\
\ForEach{point $p_{i,j}$}{
  \eIf{$p_{i,j}$ is not a \textit{change} point}{
    \eIf{$d(p_{i,j}, \mathcal{L}_{s})>t_{p2line}$}{
      Label $p_{i,j}$ as \textit{obstacle} point: $l_{i,j} = l^o$
    }{
      Label $p_{i,j}$ as \textit{ground} point: $l_{i,j} = l^g$
    }
  }
  {
    Label $p_{i,j}$ as \textit{obstacle}: $l_{i,j} = l^o$
  }
} 

\caption{Coarse to fine ground segmentation}
\label{alg::ground_segmentation}
\end{algorithm}

\textit{Fine ground segmentation:} 
After the above step, \textit{change} and \textit{ground} points have high probabilities to be truly from obstacles or ground respectively. The \textit{uncertain} points and \textit{change follow} points need more investigations. In this step, 2D line segments fitting are utilized to refine the coarse ground segmentation. The points within one packet buffer are projected into 2D plane $z-\rho^{xy}$ for line fitting. Although these points are not in the same plane in reality, this is still a good approximation. This 2D approximation allows a faster line fitting, comparing to the 3D cases as in \cite{Dim2017}. 2D line segments in $z-\rho^{xy}$ plane are: 
\begin{equation}
\mathcal{L}_s: z = a_s\rho^{xy}+b_s,\; \rho^{xy}\in[\rho_{0,s}^{xy},\rho_{1,s}^{xy}], s\in[1,S], 
\end{equation}
Where $S$ is the total number of line segments in one block, $a_s$ and $b_s$ are slope and intercept, $\rho_{0,s}^{xy},\rho_{1,s}^{xy}$ are the start and end of the $s$th line segment. 

Line segments are fitted sequentially \cite{line2007} from the \textit{ground} points or \textit{uncertain} points. To assure the line segments are flat, the following results are valid only if $a_s$ and $b_s$ are within certain ranges. Then, distances from all points except the \textit{change} points to the nearest line segment are calculated. If a distance exceeds a certain threshold $t_{p2line}$, the point is labeled as \textit{Obstacle}. Otherwise, it is classified as \textit{ground}. Notice that all the \textit{change} points are directly labeled as \textit{obstacle} points. 

\subsubsection{Clustering}
\label{sec::clustering}
\begin{algorithm}[!hbpt]
   \small
  \SetAlgoLined
  \KwIn{Obstacle points $p\{l^o\}$ after ground segmentation}
  \KwOut{Clusters $\mathcal{C}$}
\tcc{\small{Step 1: initial CCL based clustering}}
\For{every $p\{l^{o}\}$ in packet buffer}{
  \eIf{ it is the first point to process}{
    Create a new initial cluster $\tilde{\mathcal{C}}(p)$, $p\in \tilde{\mathcal{C}}(p)$}
  {
  \For {every $p'\{l^{o}\}$ in searching region $\mathcal{S}(p\{l^{o}\})$}{
    Calculate: $\Delta\rho^{xy}$\\
    \If{$\Delta\rho^{xy} < t_{\Delta\rho^{xy}}^{ccl}$}{
      Push $p$ into $\tilde{\mathcal{C}}(p')$, where $p'\in \tilde{\mathcal{C}}_k$. \;
      \textbf{break};
    }
  }
  \tcc{\small{$p\{l^o\}$ isn't assigned an cluster}}
  Create a new cluster $\tilde{\mathcal{C}}(p)$, insert it into cluster buffer\; 
}
  }

\tcc{\small{Step 2: cluster refinement}}
\For{All initial clusters $\tilde{\mathcal{C}_i}$ within cluster buffer}{
  \If{$\tilde{\mathcal{C}_i}$ has a linked cluster $\tilde{\mathcal{C}_j}$ through \textit{contain, overlap, neighbour}}{
    Calculate mutual distance: $d(\tilde{\mathcal{C}_i}, \tilde{\mathcal{C}_j})$\;
    \If{$d(\tilde{\mathcal{C}_i}, \tilde{\mathcal{C}_j}) < t_{merge}$}{
      Merge $\tilde{\mathcal{C}_i}$ into $\tilde{\mathcal{C}_j}$\;
      Update cluster information\;
    }
  }
}
\caption{Clustering non-ground points}
\label{alg::clustering}
\end{algorithm}

We use connected component labeling (CCL) to quickly group points into initial clusters $\widetilde{\mathcal{C}}$. From left to right, the algorithm sweeps each column from bottom to top. Given a point $p$, its search region $\mathcal{S}(p)$ is defined as previous $N$ columns along the sweeping order. Its connectivity to a point $p'$ within $\mathcal{S}(p)$ is decided by evaluating range difference $\Delta\rho^{xy}$. 
Although $\Delta\rho^{xy}(p,p')$ is not the Cartesian distance, we found it works very well when $\mathcal{S}(p)$ is not big. If $\Delta\rho^{xy}(p,p')$ is smaller than a threshold $t_{\Delta\rho^{xy}}^{ccl}$, $p$ and $p'$ are connected and grouped into the same cluster. Otherwise, if all the points within $\mathcal{S}(p)$ are not connected to $p$, a new cluster starting from $p$ is created. $\mathcal{S}(p)$ is usually defined as several previous columns, i.e. 5, in received packet buffers. This method is described in Algorithm \ref{alg::clustering}.

\textit{Refinement:}
The initial clustering can quickly group adjacent points together. However, it leaves the problem of over-segmentation untouched. The main cause of over-segmentation is the various reflectivities of different parts in one object. For example, due to windshield is transparent for LiDAR, vehicle's window parts are easily over-segmented. Another common reason is the occlusion caused by small objects, such as a pole on the roadside. 

\begin{figure}[!tbhp]
  \centering
\includegraphics[width = 0.25\textwidth]{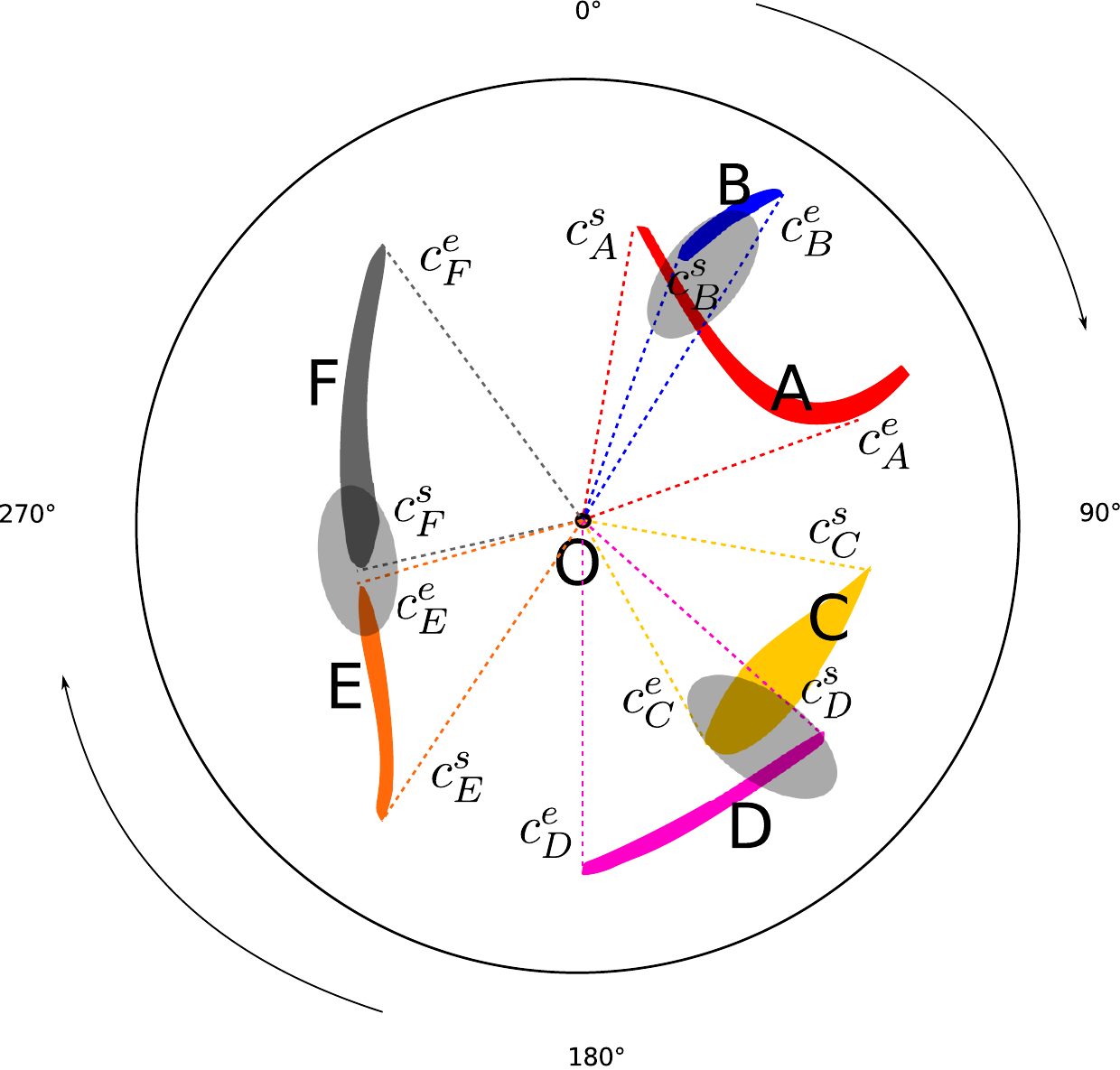}
\caption{Clustering refinement examples: A contains B, C overlaps D, E and F are neighbours.}
\label{fig::clustering}
\end{figure}

To find potentially over-segmented clusters, we explore the linkages of initial clusters $\widetilde{\mathcal{C}}$, based on their positions in polar coordinates. Given $\widetilde{\mathcal{C}_1}$ and $\widetilde{\mathcal{C}_2}$, they may be linked by \textit{contain, overlap} or \textit{neighbour} relation, as demonstrated in Fig .\ref{fig::clustering}. The conditions of these three linkages are defined as:
\begin{itemize}
\item \textit{Contain}: $[\tilde{c}_2^s, \tilde{c}_2^e]\in[\tilde{c}_1^s,\tilde{c}_1^e]$ or $[\tilde{c}_1^s, \tilde{c}_1^e]\in[\tilde{c}_2^s,\tilde{c}_2^e]$
\item \textit{Overlap}: $\tilde{c}_1^s\in[\tilde{c}_2^s,\tilde{c}_2^e], \tilde{c}_2^s\notin [\tilde{c_2^s},\tilde{c}_2^e]$ or  vise versa. 
\item \textit{Neighbour}: $gap(\widetilde{\mathcal{C}}_1, \widetilde{\mathcal{C}}_2)< t_{neighbour}$, where $gap(*)$ computes the number of gap columns between two clusters.
\end{itemize}

The linked initial clusters are merged if they are close enough. The distance between two clusters is defined as the $n$th minimum mutual point Cartesian distances within search regions respectively:
\begin{equation}
  d(\tilde{\mathcal{C}}_1, \tilde{\mathcal{C}}_2) = \min_{n}d(p,p'), p\in \mathcal{S}(\tilde{\mathcal{C}}_1|\tilde{\mathcal{C}}_2), p'\in \mathcal{S}(\tilde{\mathcal{C}}_2|\tilde{\mathcal{C}}_1)
  \label{eq::cluster_distance}
\end{equation}
The search region is defined as the overlapping area or adjacent parts, as shown in Fig.\ref{fig::clustering}. 


\section{Experiments}\label{sec::experiment}
The proposed method is implemented in C++ under ROS Kinetic framework, within a PC with an Intel CPU i7-7820, 16G memory. A Velodyne VLP32C is mounted on the roof of a Renault ZOE. We modified an open-source Velodyne driver to directly output raw data packets. 
The proposed method is compared with two other open-source methods: (1) Scan Line Run (SLR) \cite{Dim2017}  and (2) DepthClustering (DC) \cite{Igor2016}).


\subsection{Evaluation Metrics}
To have quantitative evaluations, similar to \cite{Shin2017}, we used \textit{precision, recall}, \textit{TPR (true positive rate)}, \textit{FNR (false negative rate)}, \textit{OSR (Over-segmentation suppression rate)} and \textit{USR (Under-segmentation suppression rate)} as metrics. TPR, FNR, OSR and USR are defined in Eq. \ref{eq::metrics}.
\begin{equation}
  \begin{split}
    &TPR = TPs/All, FNR = FNs/All\\
    &OSR = TPs/(TPs + \textit{over-segmentation)}\\
    &USR = TPs/(TPs + \textit{under-segmentation})
  \end{split}
  \label{eq::metrics}
\end{equation}
where $All$ is the number of total "true objects" (usually acquired by annotation). True Positives (\textit{TPs}) are the detections correctly segmented. False Positives (\textit{FPs}) are the phantom detections. False Negatives (\textit{FNs}) are the missed detections. \textit{Over-segmentation} means an object is segmented into several ones. In contrast, \textit{under-segmentation} is the case of merging several objects into one segmentation. In general, higher $OSR$ or $USR$ means less \textit{over-segmentations} or \textit{under-segmentation}, which is preferred.


\begin{figure}[t]
\centering
\subfigure[Over-segmentation sample]{
\includegraphics[width = 0.2\textwidth, height = 2.8cm]{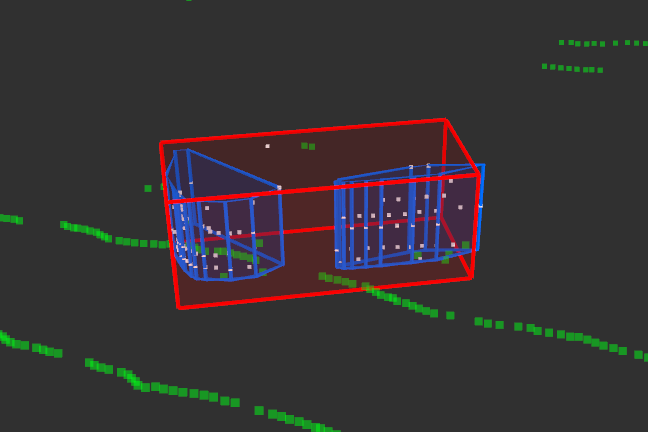}
}
\subfigure[Under-segmentation sample]{
\includegraphics[width = 0.2\textwidth, height = 2.8cm]{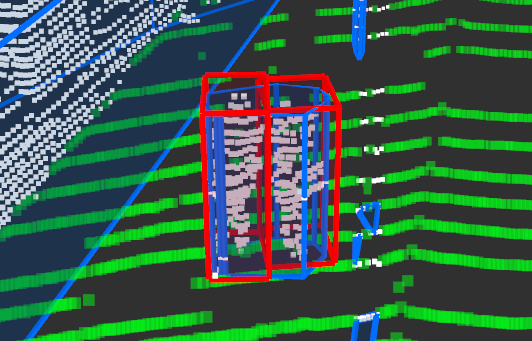}
}
\caption{Examples of over-segmentation, under-segmentation used for quality evaluation.}
\label{fig::4cases}
\end{figure}

\subsection{Test in Rouen dataset}
The proposed methods are evaluated for a dataset recorded by our platform, in the center of Rouen, France. The recorded dataset contains 20 minutes around 12K frames of point clouds. We annotate all the objects being vertical to the ground. 4860 objects from 200 frames within 30 meters are manually extracted. This dataset also contains raw LiDAR data packets that enable the test of the proposed method.


\subsubsection{Parameter settings}
The range images sizes for SLR and our method are $32*1800$, for depth clustering, the size is $32*900$. The parameters of SLR are tuned as: $N_{seg}=4, N_{iter}=5, N_{LPR} = 20, Th_{seeds} = 0.5m$ $Th_{dist}=0.2m, Th_{run}=0.3m$ and $Th_{merge}=1m$, which are slightly different to the settings in \cite{Dim2017}. For depth clustering, the parameters are optimized as: the window size for Savitsky-Golay smoothing: 5, ground removal angle threshold: $8^\circ$, clustering angle tolerance is $11^\circ$.

The packet buffer $\mathcal{B}$ is set to 5 data packets. Other parameters are: $t_{\alpha} = 0.5$, $t_{\Delta\rho^{xy}} = 2m$. For fine ground segmentation, the block size $B = 20$, and $t_{p2line} = 0.2m$. In initial clustering, $t_{\Delta\rho^{xy}}^{ccl}$ is $1m$. For the search region $\mathcal{S}(p)$, if $\mathcal{S}(p)$'s range is less than 20m, $\mathcal{S}(p)$ is the previous 5 columns. Otherwise, $\mathcal{S}(p)$ is its previous 10 columns in the range image. In the refinement step, $t_{neighbour}$, $d(\tilde{\mathcal{C}}_i, \tilde{\mathcal{C}}_j)$, $t_{merge}$ are set to 5, the third minimum mutual point distance, and $0.8m$ respectively.

\begin{table}[t]
  \centering
  \renewcommand{\arraystretch}{1.1}
  \footnotesize
\begin{tabular}{r c c c}
\toprule
   \multicolumn{4}{c}{\textbf{Rouen dataset: 200 frames, 4860 objects}} \\\midrule           
   &Our & SLR & \tabincell{c}{Depth \\Clustering} \\\midrule
  \textit{Precision} &\textbf{0.987}  &0.968  &0.952  \\
  \textit{Recall} &\textbf{0.991}  &0.987  &\textbf{0.990}  \\
  \textit{TPR} &\textbf{0.917}  &0.892  &0.861  \\
  \textit{FNR} &0.008  &0.010 &\textbf{0.007}  \\
  \textit{USR} & 0.982  &0.985  &\textbf{0.988}  \\
  \textit{OSR} &\textbf{0.963}  &0.914 &0.831  \\
  \bottomrule
\end{tabular}
\vspace{0.2cm}
\caption{Quantitative evaluation of three methods in Rouen dataset.}
\label{tab::evaluation}
\end{table}

\begin{figure*}[t]
  \centering
\subfigure[Rouen dataset example]{
\includegraphics[width = 0.31\textwidth,  height=4cm]{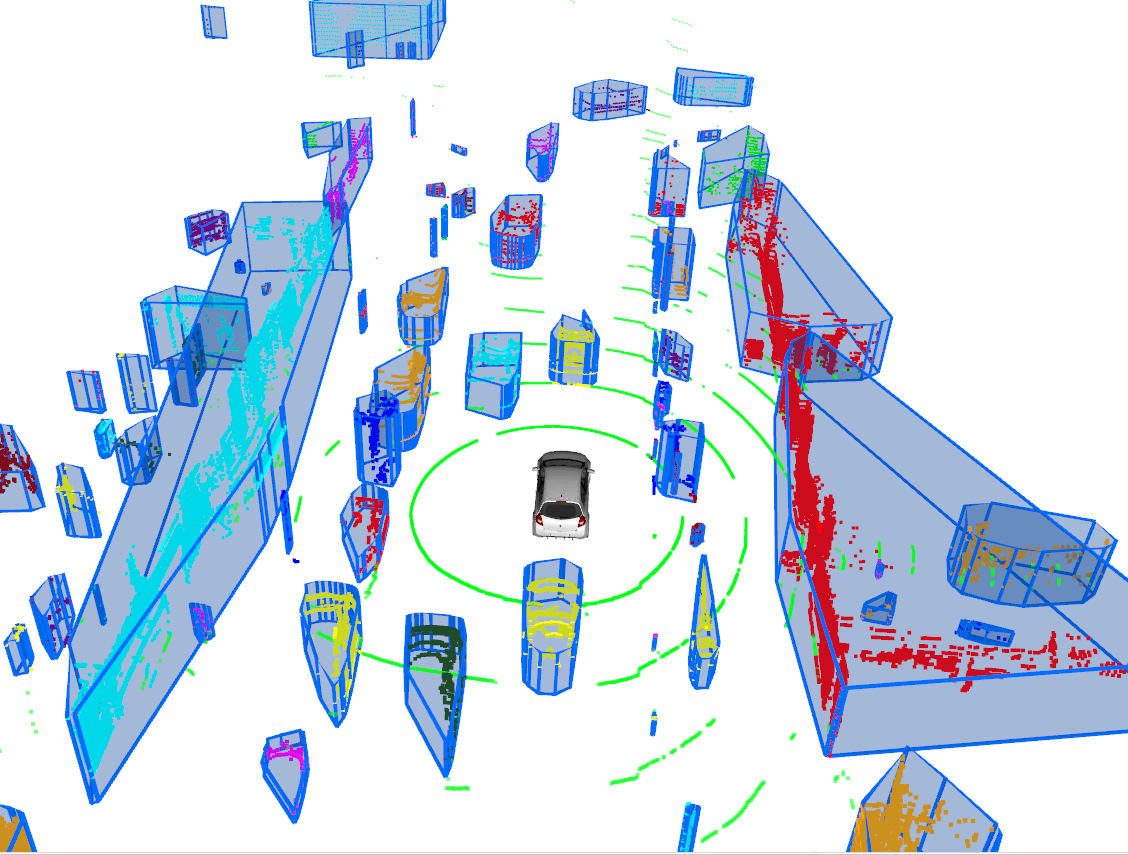}
}
\subfigure[Rouen dataset example]{
\includegraphics[width = 0.31\textwidth,  height=4cm]{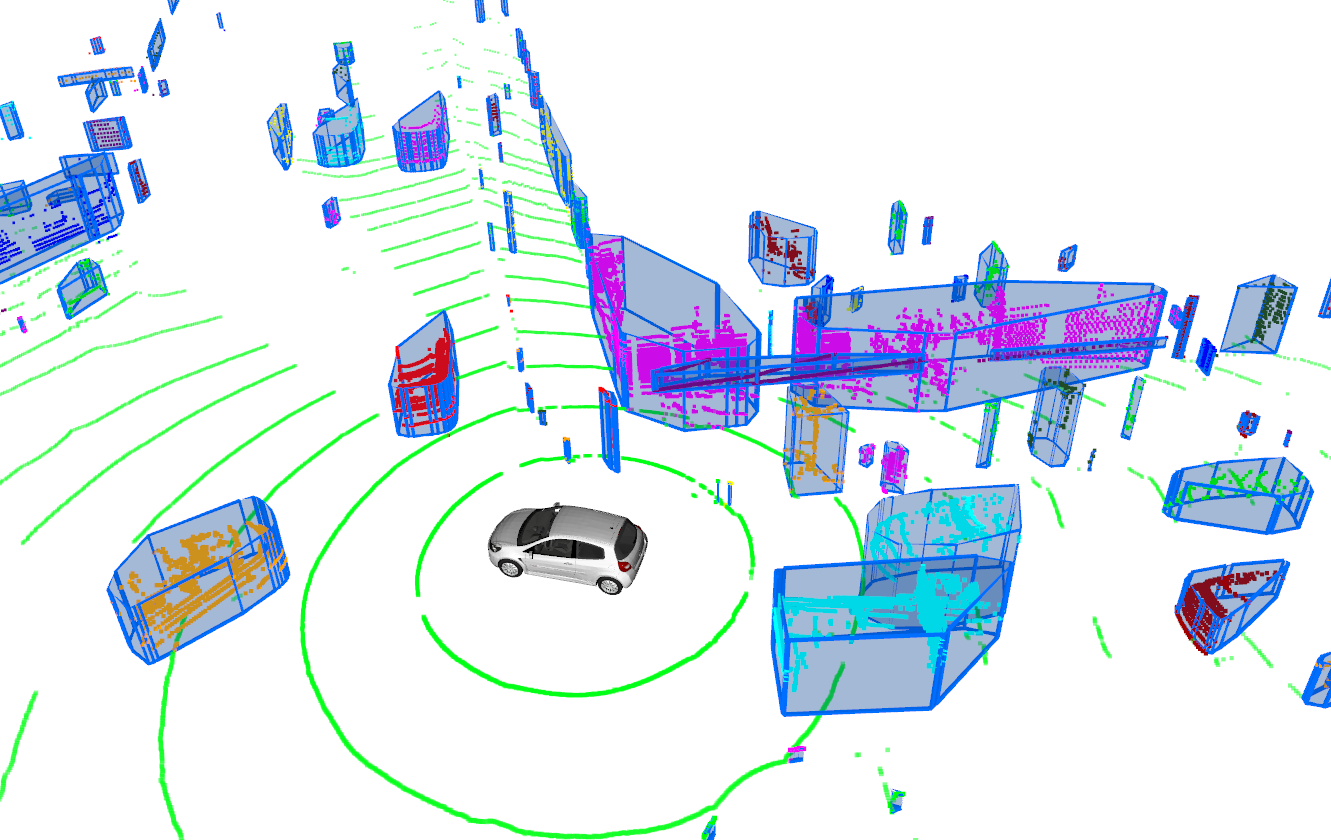}
}
\subfigure[Rouen dataset example]{
  \includegraphics[width = 0.31\textwidth,  height=4cm]{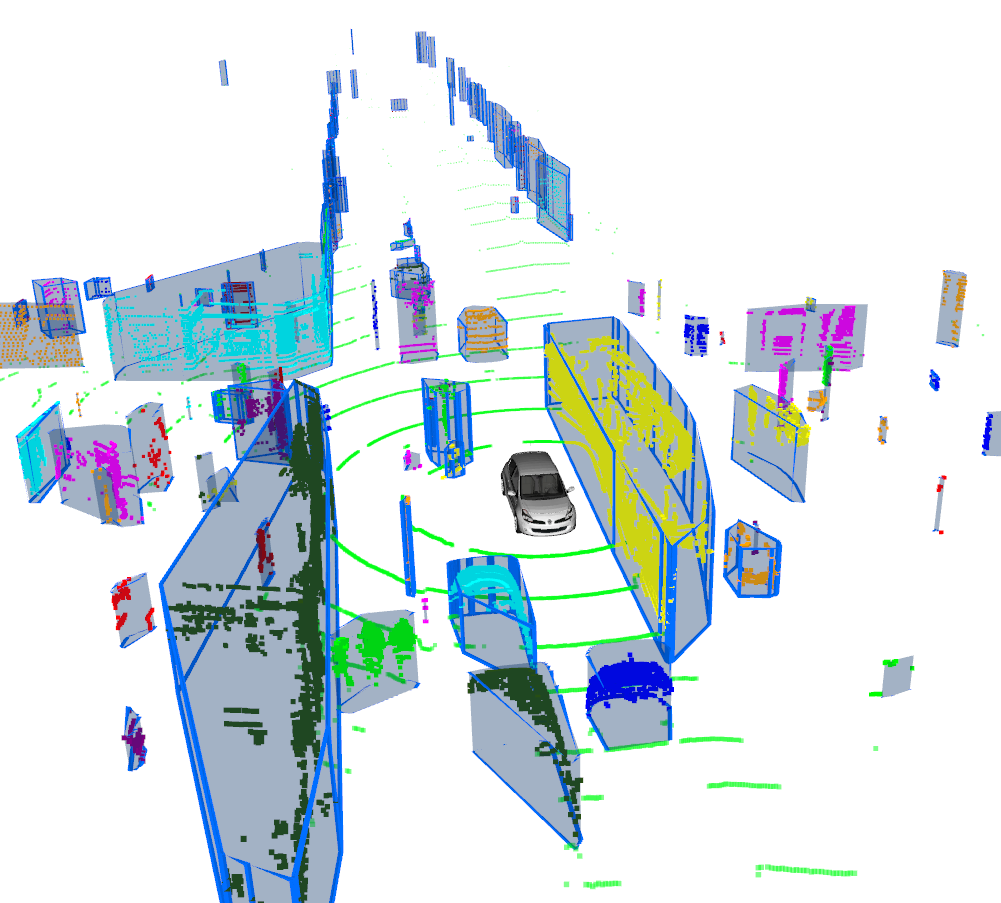}
}
\subfigure[Failure example 1: under-segmentation]{
\includegraphics[width = 0.22\textwidth, height = 3cm]{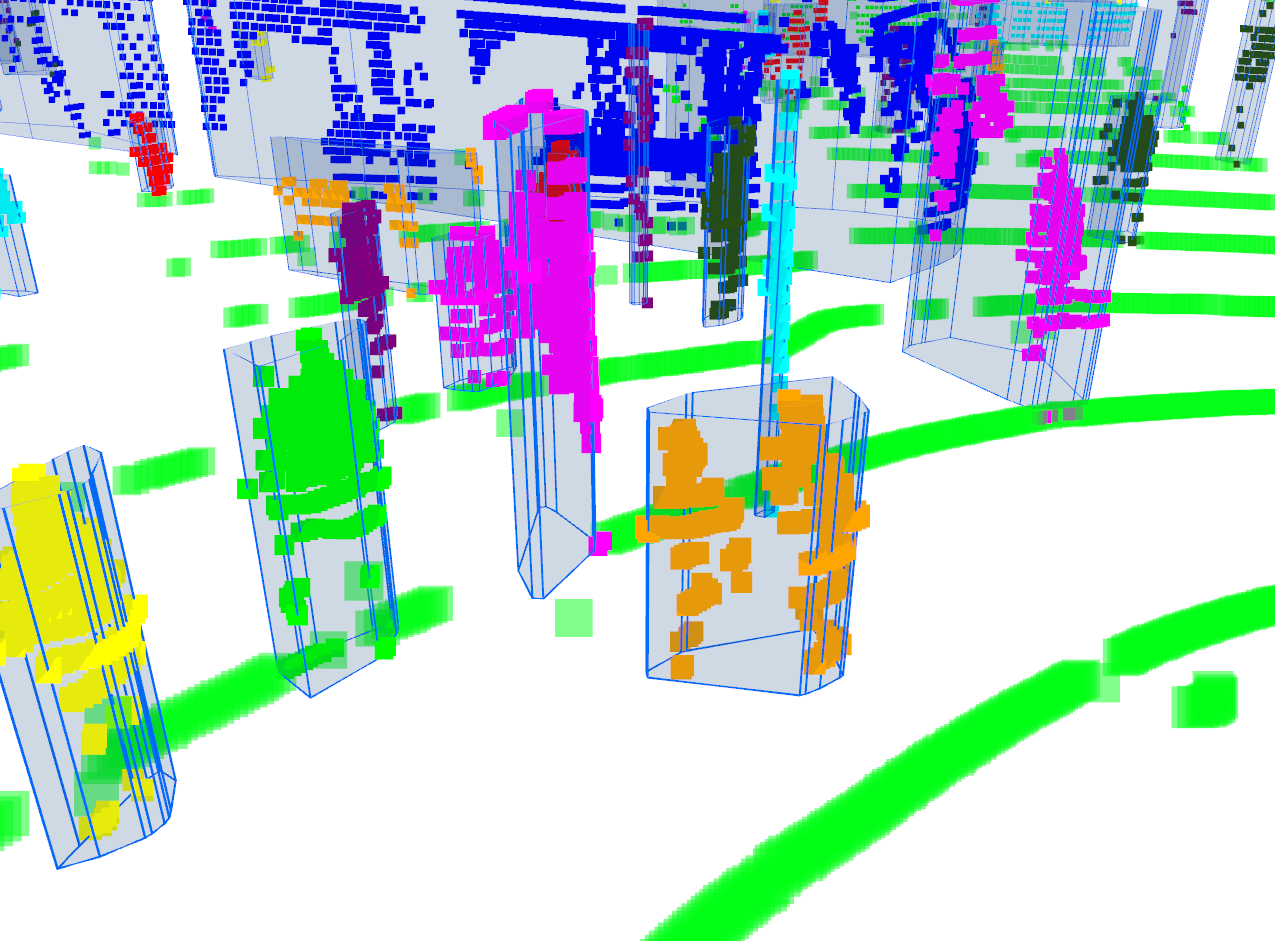}
}
\subfigure[Failure example 2: under-segmentation]{
\includegraphics[width = 0.22\textwidth, height = 3cm]{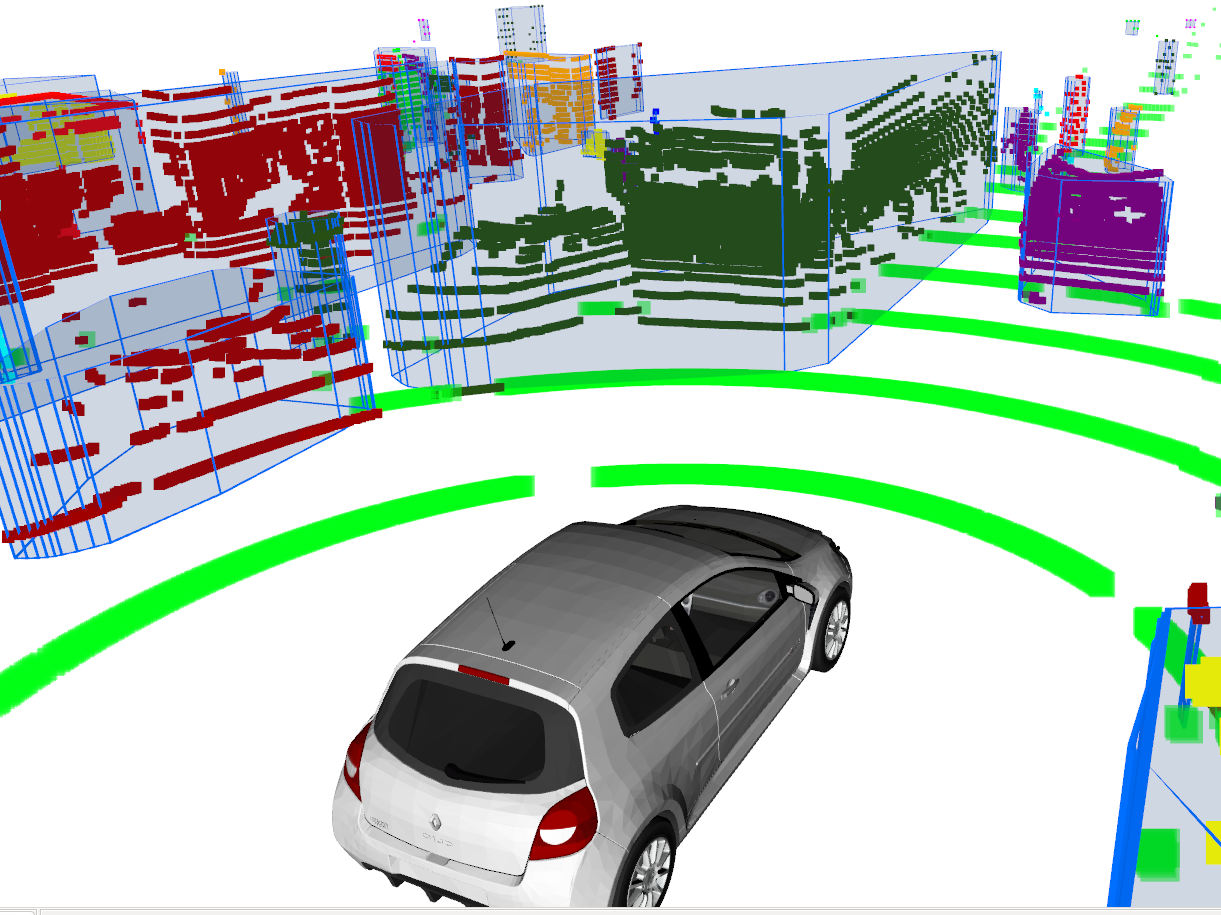}
}
\subfigure[Failure example 3: over-segmentation]{
\includegraphics[width = 0.22\textwidth, height = 3cm]{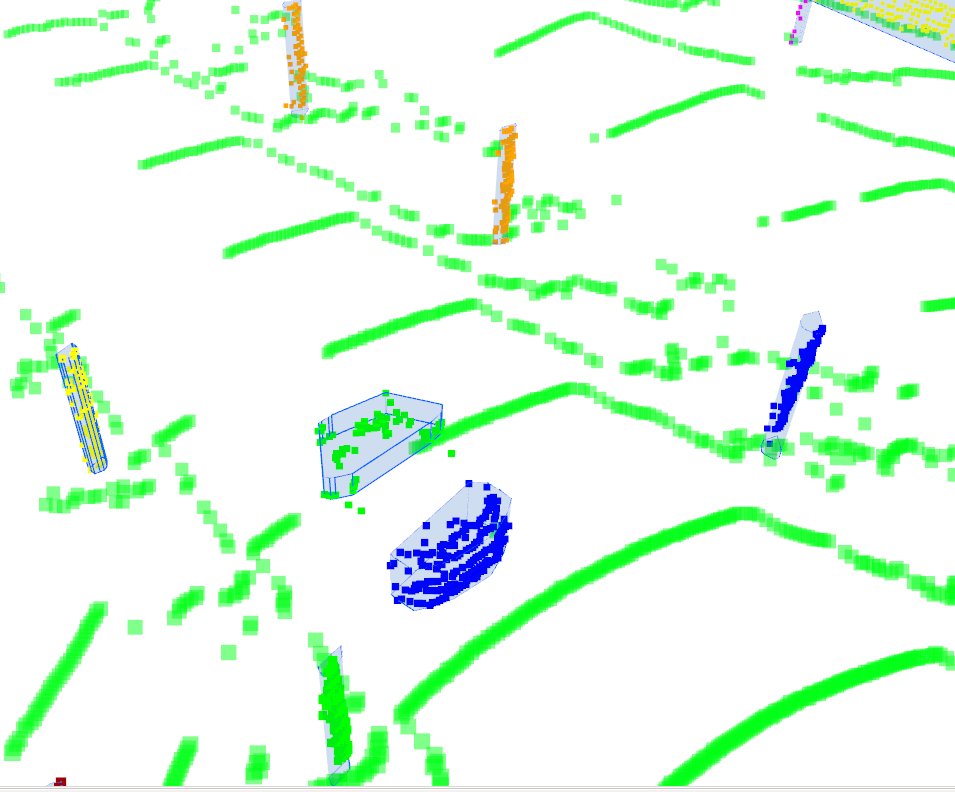}
}
\subfigure[Failure example 4: over-segmentation]{
\includegraphics[width = 0.22\textwidth, height = 3cm]{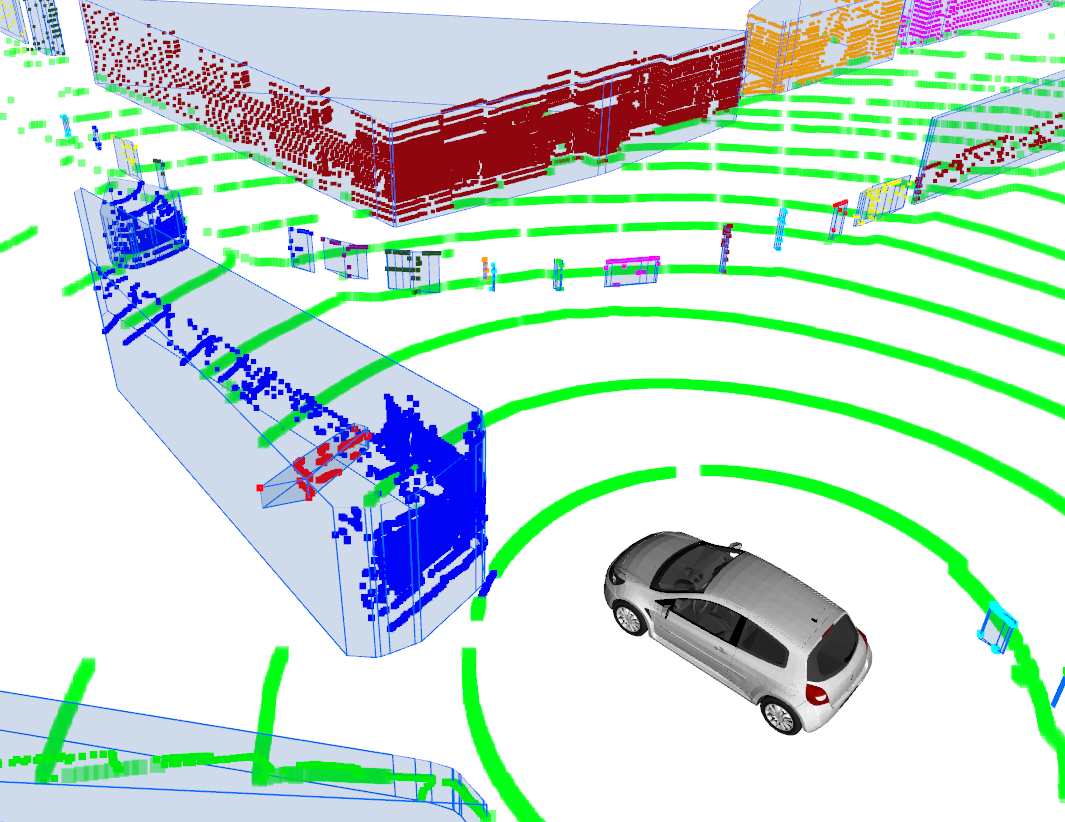}
}
\caption{More results of our segmentation method on Rouen dataset.}
\label{fig::more_results}
\end{figure*}

\subsubsection{Results analysis}
From the results shown in Tab. \ref{tab::evaluation}, our method achieves the best scores in \textit{Precision, Recall}, and \textit{TPR}. Depth clustering gets the least \textit{FNR}. Our method is the best in OSR, which means less in over-segmentation. However, the price is the lowest USR, which means our method is apt to merge nearby objects. In opposite, depth clustering has the highest USR score while the lowest OSR that this method tends to split objects. SLR has balanced performances in all the metrics.As for the processing speed (in Tab. \ref{tab::processing_time}), InsClustering is much faster than other methods that are based on range image. The reason is that, by processing directly on raw data flow, we separate the total processing time for a whole range image into many subsets. Hence, when the LiDAR finishes a circle of scanning, the processing of our method (raw data version) is almost finished. While for a range image based method, it is just the beginning. 

\begin{table}[t]
  \centering
  \footnotesize
  \renewcommand{\arraystretch}{1.1}
\begin{tabular}{r p{0.7cm} p{0.7cm} p{0.7cm}}
\toprule
   &\textbf{All} &\tabincell{c}{Ground\\filtering} &\tabincell{c}{Cluster\\-ing}\\\hline
  SLR &\scriptsize{30.8ms} &\scriptsize{13.8ms} &\scriptsize{17.0ms} \\\hline
  \tabincell{r}{Depth\\Clustering} & \scriptsize{5.5ms} &\scriptsize{2.3ms} &\scriptsize{3.2ms} \\\hline
  \tabincell{r}{Proposed\\InsClustering}  &\scriptsize{\textbf{265us}}  &\scriptsize{98us} &\scriptsize{167us}\\
  \bottomrule
\end{tabular}
\vspace{0.2cm}
\caption{Benchmarking average processing time}
\label{tab::processing_time}
\end{table}


\subsubsection{Qualitative Examples}
To directly understand the performance of the proposed method, we qualitatively demonstrate the segmentation results in Rouen dataset, as shown in Fig. \ref{fig::more_results}, where some failure examples for over-segmentation and under-segmentation are demonstrated.

\section{Conclusions}\label{sec::conclusion}
In this paper, we propose the InsClustering: a fast point cloud clustering method designed for processing raw data flow from spinning LiDAR. Thanks to this design, our method is proved much faster than two SOTA methods. Meanwhile, a coarse-to-fine framework is adopted in both ground segmentation and object clustering. Experimental results show the fast speed and good accuracy.

\bibliographystyle{IEEEtran}
\bibliography{IEEEabrv,ms}
\end{document}